# Emergent effects of scaling on the functional hierarchies within large language models

Paul C. Bogdan (paulcbogdan@gmail.com)
Duke University

## Extended Abstract

**Introduction:** Transformer architectures usually involve repeated transformer layers of the same size. However, the architecture is often described as functionally hierarchical: Early layers process syntax, middle layers begin to parse semantics, and late layers integrate information across a wide scale. The present work revisits these ideas within large language models (LLMs).

**Methods:** The present research submits simple texts to an LLM (e.g., "A church and organ"), and resulting activations at different layers are extracted. Then, for each layer, support vector machines (SVM) and ridge regressions are fit. The classifiers attempt to predict some property of the texts based on a layer's activations – e.g., (i) whether an object has some semantic feature, (ii) whether two objects are related, or (iii) whether a four-object analogy is valid. High accuracy indicates that a label's information is expressed within a given LLM layer. To understand more global processing and how information may percolate through layers in longer texts, experiments were also conducted while appending ~100 words of filler to the end of the short target texts. This general approach was used with contemporary models that are either small (Llama-3.2-3b; 28 layers) or large (Llama-3.3-70b-Instruct; 80 layers).

**Results:** The small-model analyses partly bolster the common hierarchical perspective: In short texts, item-level semantics are most strongly represented early (layers 2-7), then two-item relations (layers 8-12), and then four-item analogies (layers 10-15). Afterward, the representation of items and simple relations gradually decreases in deeper layers that focus on more global information, and these patterns are overall consistent with a gradual abstraction hierarchy.

However, there also arise several findings that run counter to a steady hierarchy view: *First*, although deep layers can represent document-wide abstractions, deep layers also represent narrow two-item relations from historically back in a context window. That is, deep layers can broadly compress information, potentially without meaningful abstraction. *Second*, when examining the larger 80-layer Llama model, there appear to be stark fluctuations in abstraction level: As depth increases, two-item relations and analogies initially increase in their representation (peaking layers 12-16), markedly decrease, and then increase again momentarily (peaking again in layers 25-33). This peculiar pattern consistently emerges across several experiments of two-item relations or four-object analogies. Thus, models seem to fluctuate in their level of abstraction, and/or there may exist multiple parallel abstraction hierarchies at play in large models. *Third*, another emergent property of scaling is coordination between the attention mechanisms of adjacent layers. Consistently across multiple experiments, adjacent layers appear to fluctuate between what information they each specialize in representing, and this only occurs in the large 70b model.

**Conclusion:** These results shed light on the functional architecture and the emergent organization of transformer networks: An abstraction hierarchy often manifests across layers, but large models also deviate from this pattern in curious ways. Along with these findings, the layer-wise map laid out will hopefully inform work extracting embeddings from LLM or attempting to steer them through activation engineering.

## Introduction

Modern LLMs are generally based on the transformer architecture, consisting of identical transformer layers stacked in sequence. Despite layer size being constant, many researchers hypothesize that these layers develop a functional hierarchy during training: Early layers process syntax, middle layers begin to handle semantics, and deeper layers integrate information more broadly.[1,2] This hypothesized hierarchical structure is pervasive in descriptions of LLM functioning, and it presumably stems from the prevalence of abstraction hierarchies across machine learning models on different domains.

Yet empirical investigation of this posed functional hierarchy remains limited, particularly for modern LLMs. Most studies providing layer-by-layer analyses have instead relied on older LLMs (often BERT), which may not reflect the emergent properties of contemporary models.[2–4] In addition, prior studies have usually focused on carefully understanding just one form of information – e.g., sentence-level semantic similarity – while not actually mapping out a gradient of semantic abstraction.[5]

To investigate functional hierarchies in modern LLMs and the roles of different layers, I conducted several experiments. Each experiment follows a common structure: A dataset of tiny or medium-length texts was prepared (e.g., "An apple"). Each text was submitted to an LLM (Llama-3.2-3b or Llama-3.3-70b-Instruct). The LLMs' resulting activation patterns were extracted, separately for each layer. These activations were submitted to support vector machines (SVMs) or ridge regressions, attempting to predict some property of the input text (e.g., whether the object in "An apple" is edible). By varying the input text and its labels, my analyses attempt to tease out the information encoded (linearly) at different levels.

## Experiment 1: Item-level semantics

Analyses first investigated item-level semantic coding. This experiment referenced public data from a human study on semantic processing (https://cslb.psychol.cam.ac.uk/propnorms). In the study, participants were shown objects and asked to come up with features describing each one (e.g., when shown an "apple," a participant may respond "is edible" or "is a fruit"). The present analyses were based on 20 of the most common features and 300 of the objects associated with the most features.

For each object, Llama-3.2-3b was fed a two-word text (e.g., "An apple") and activations corresponding to the object were extracted; for words split across multiple tokens, the tokens were averaged; loose analyses suggested this was better than taking the activations of just the last token. Activations were extracted separately for (i) the residual stream input into each transformer layer, (ii) the outputs of the attention layer added to the stream, e.g., the outputs added to the "apple" tokens based on all prior tokens, and (iii) the outputs of the feed-forward network layer added to the stream. Separately for each of these components each layer, SVMs were fit for each of the 20 human-reported features. Each SVM attempted to predict whether an object had (1) or did not have (0) each feature, using standard 6-fold cross-validation. **Figure 1a** shows the achieved accuracies averaged across all twenty item features, and **Figures 1b-d** show the accuracies achieved for each item feature, plotted separately for the three parts of the transformer layer probed.

Three potentially interesting patterns emerge. *First*, the representation of item-level semantics in the residual stream and feed-forward network (FFN) output peaks in early layers (layers 2-7) before descending but turning upwards again in the final layers. This rise, fall, and ultimate brief rise have been shown in other layer-by-layer studies on LLMs.[4,6] This initial rise and then fall may reflect the layers steadily increasing in their level of abstraction, such that the peak describes where item-level semantics exist along this abstraction gradient. Then, the shift and final ascent may reflect a decrease in abstraction for next-word prediction. *Second*, despite examining singular items, representation does not peak in the input to the first layer, which corresponds to the input embeddings. Those embeddings thus may be structured in a way specifically to process syntax rather than semantics, or this layer-by-layer warm-up period may be linked to multi-token words becoming unified (e.g., if " apple" is tokenized as " app" and "le", then the initial layers unify this into an apple representation). *Third*, the attention outputs represent item-level semantics despite the absence of meaningful relationships between words. This likely reflects how the residual stream's representation is expressed somewhat in the query vector of the object. The first point here, on the abstraction hierarchy, would be further investigated with experiments exploring more abstract information, attempting to stimulate coding in deeper layers.

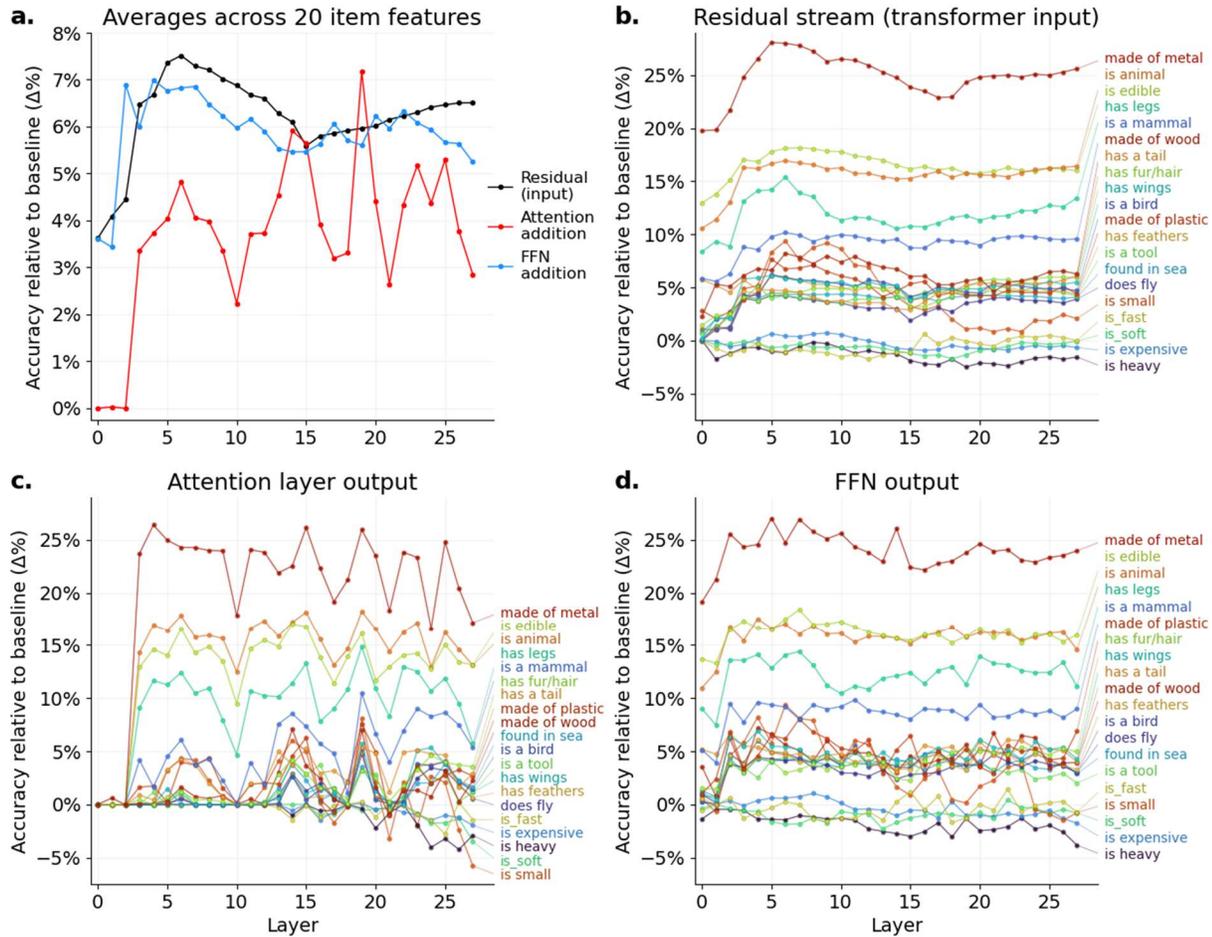

*Figure 1. Item-level representation. Accuracies for the binary prediction tests of Experiment 1 on Llama-3.2-3b. **a.** Achieved accuracies averaged across all twenty item features. **b-d.** Accuracies achieved for each item feature, plotted separately for three components of a transformer layer.*

**Experiment 2: Two-item relations**

The experiment 2 analyses investigated how transformers encode the relations between two items, across three branches of tests. The tests examined how an LLM encodes: (A) whether an object is likely to be found in a scene, (B) broadly, whether two items are semantically related in some way, and (C) more narrowly, whether a herbivore/carnivore would eat a plant/meat. Each of these tests followed a similar structure as in Experiment 1, where texts are crafted and submitted to Llama-3.2-3b, and then the resulting activations of different layers are used to fit classifiers.

**Experiment 2A**

This experiment examined how a model encodes object-scene relationships and used data from a study by my research group. In the study, participants were shown 114 objects, each paired with one of 114 scenes. For example, participants may be shown a *tractor* on a *farm* (likely to be found), a *bug* in a *desert* (moderately likely), or a *poker table* in a *prison* (unlikely to be found). Overall, 342 object-scene pairs were designed and administered to participants. For each object-scene pair, participants used a 4-point scale to report how likely they perceived it to be for the object to be found in the scene (1 = "Very unlikely"; 4 = "Very likely"). For the present LLM research, each object-scene pair's likeliness score was averaged across all participants, and I investigated how the LLM encodes this likeliness.

The object-scene pairs were used to produce texts ("In a {SCENE}, an {OBJECT}"), which were submitted to Llama-3.2-3b, and activations were extracted corresponding to the object embedding. The activations were used to fit Ridge regressions predicting the object-scene's likeliness score. Group-6-fold cross-validation was used such that scene-object examples were grouped by object, and these groups were preserved across the folds. **Figure 2a (left)** shows the cross-validated accuracies. The representation of the relation in the residual stream peaked at layer 11 – noticeably deeper than the item-level semantics peak in Figure 1. This analysis was also done while defining the texts as "A {SCENE} and {OBJECT}", which yielded similar patterns (**Figure 2a right**). Interestingly, there was less of a dip following this peak, which may reflect scene/context words having a type of global relevance represented in deep layers, but this was not otherwise tested. Most relevant though for the present work was this two-item semantic relation peak being deeper than the one-item semantics earlier. This is consistent with model layers progressively increasing in abstraction.

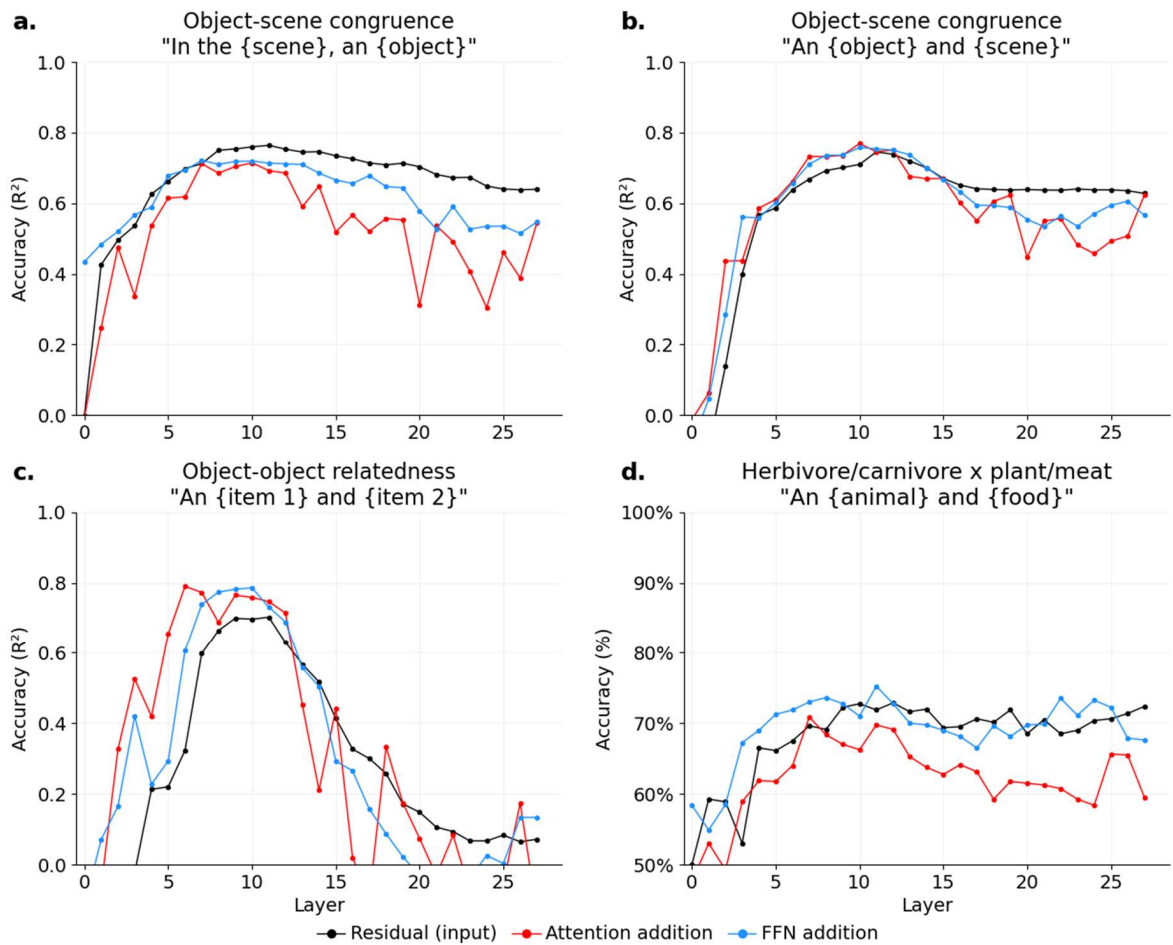

*Figure 2. Two-item-relation representation.* Accuracies associated with the three Llama-3.2-3b experiments specifically on two-item relations. The subfigure letters correspond to Experiments 2A, 2B, and 2C; the results from the two forms of Experiment 2A are reported separately on the left and right sides.

### Experiment 2B

This next experiment examined two-item relations between pairs of objects. The experiment leveraged human data from a study where participants used a 4-point scale to rate the relatedness words in pairs (e.g., "church" and "organ") (https://osf.io/5q6th/). Participants rated this for sixty word pairs, which were used here to produce the texts "An {ITEM 1} and {ITEM 2}" and the flipped "An {ITEM 2} and {ITEM 1}". These texts were submitted to Llama-3.2-3b. Activations were extracted and then submitted to an SVM, which yielded the accuracies shown in Figure 2b. The peak is consistent with those of Figure 2a. Interestingly, the drop in later layers is much starker, although this result overall provides more evidence that two-item relation representation peak proceeds the one-item semantic representation peak.

### Experiment 2C

A final test of two-item relations studied this topic more narrowly than the type of general "relatedness" examined thus far. Here, I prepared two-item texts where one item was always a food (plant or meat) and the other was an animal (herbivore or carnivore) (e.g., "A {FOOD} and {ANIMAL}"). The texts were submitted to Llama-3.2-3b, activations were extracted, and SVMs were fit, predicting whether the text's animal would eat the food. Cross-validation was structured using 2-group folds, where training was always done on herbivores then testing on carnivores or vice versa; this required z-normalizing the activations with respect to the food examples, as each food would flip labels between the training and testing sets. This procedure was done using texts where the food was mentioned first (e.g., "A leaf and a deer") to generate accuracy estimates, and the procedure was also independently done using texts where the food was mentioned second (e.g., "A deer and a leaf") to generate accuracy estimates. The accuracies were averaged across these two versions, and those are reported in Figure 2C. Again, we see a peak layer 10, providing further evidence of two-item relations being parsed following the item-level semantic effects of Figure 1.

### Experiment 3: Four-object analogies

To reach yet deeper layers, the next analyses considered four-object analogies. Using public LLM websites (e.g., Claude and ChatGPT), I generated fifty analogies and prepared texts for them (e.g., "Like a seed and a tree, an egg and a chicken"). For each analogy (AB:CD), three analogous valid variants were produced (BA:DC, CD:AB, DC:BA). For the classification, the analogies were also used to prepare invalid variants that an SVM would distinguish. There were two types of invalid analogies: *Easy invalids* were produced where the second and fourth words were flipped (e.g., "Like a seed and a chicken, an egg and a tree"). Distinguishing these from the valid analogies is easy because it can be done either while noticing a discrepancy between just the first two-item relation, just the second, or both. *Hard invalids* were produced by flipping just the third and fourth words (e.g., "Like a seed and a tree, a chicken and an egg"); thus, both first and second halves of the text still contain two-item relations and invalidity only emerges if the two halves are contrasted analogically.

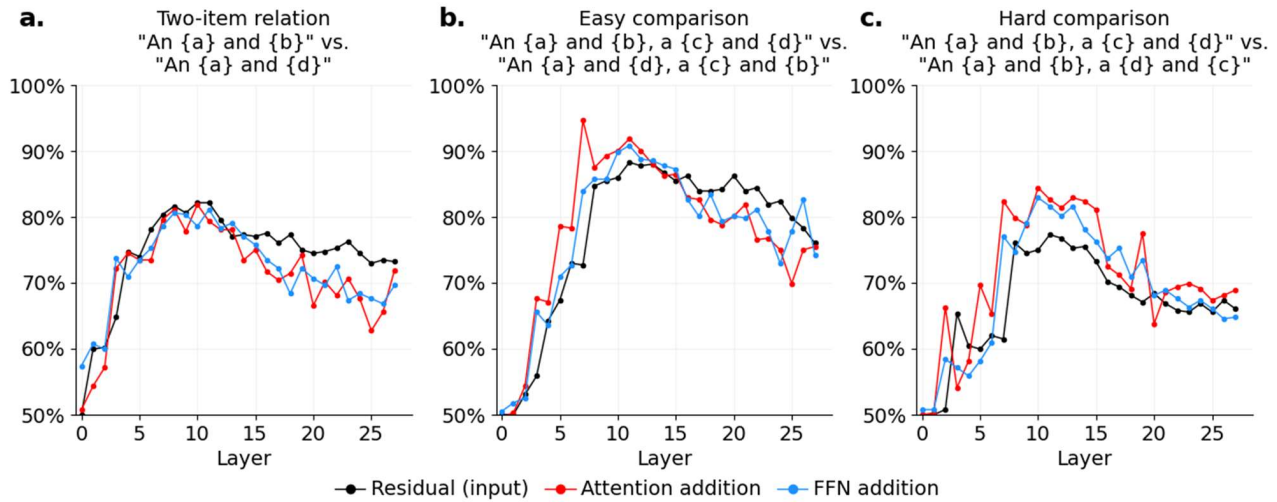

*Figure 3. Four-item-analogy representation.* Accuracies associated with the three tests specifically on four-item analogies using Llama-3.2-3b. These are all referred to as being the same "experiment" because they all use the same fifty-analogy dataset.

The accuracy patterns suggest analogical processing occurs deeper than two-item relational processing. To serve as a reference point, **Figure 3a** provides new two-item relation results using the present dataset. SVMs classified the valid versus easy invalid examples based on the activations at the second word – similar to the object-object relation classification of Experiment 2B; the peak of the two-item relation was seen around layer 10 as before. **Figure 3b** next shows the valid/easy-invalid classification results based on the four word's activations, which produced a peak slightly later. **Figure 3c** demonstrates another similarly later peak for valid/hard-invalid classification. These differences relative to two-item-relation representation are more subtle than the earlier differences with item-level representation. These patterns nonetheless are consistent with a functional abstraction hierarchy linked to layer progression. However, the next analyses would challenge a simplistic perspective of this hierarchy.

**Experiment 4: Burying text**

In an attempt to manipulate the layers in the back half of Llama-3.2-3b, the initial texts were lengthened, and classification attempted to distinguish texts based on early content. The texts were lengthened by burying the target content behind roughly 100 words of filler. This was performed for Experiments 1, 2, and 3. For example, the Experiment 2B texts, which consisted of "An {ITEM 1} and {ITEM 2}", were expanded to:

*"An {ITEM 1} and {ITEM 2} are here. I thought about this for a long while. The more I pondered, the clearer it became that my initial reaction was just the tip of the iceberg. There were layers to this issue, complexities that I hadn't considered at first glance. Each new angle brought a different perspective, challenging my assumptions and making me question what I thought I knew. It was like peeling an onion, revealing not just answers, but more questions, more nuances to explore"*

This filler

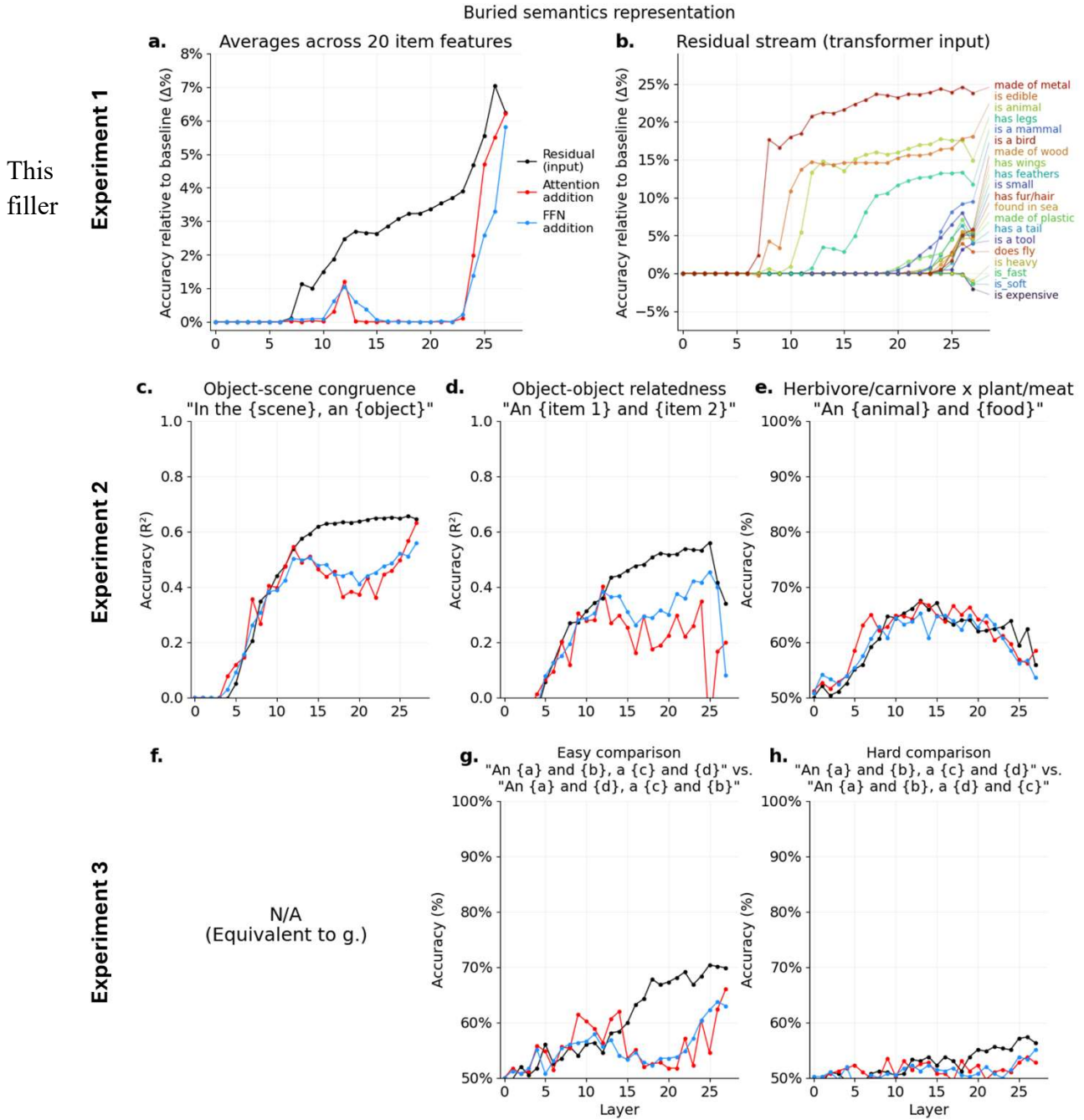

***Figure 4. Buried information representation.*** *The graphs show the results of buried-concept analyses, which are each an adaptation of an earlier experiment. Accordingly, each subfigure here maps to one of the subfigures from (top row) Figure 1, (middle row) Figure 2, or (bottom row) Figure 3. The present figure is broadly designed for a parallel structure with those earlier figures. For Experiment 3 with analogy texts, the original two-item relations experiment is equivalent to the easy analogy comparison; originally, those differed based on whether activations were extracted from the second or four words' tokens, but in the buried context, activations are always taken from the last word of the suffix filler text.*

suffix was held constant for every experiment, except for the "are here" portion at the beginning, which was changed as necessary to produce grammatically proper sentences from the start. Activations were extracted from the last word ("explore"). Thus, an SVM trained on the activations would make predictions about that initial sentence (e.g., whether items 1 and 2 are related) based on its signature that is preserved through the filler text. Despite the challenges that the filler text presumably creates, **Figure 4** shows that SVM accuracy was still considerably above chance. However, representations are consistently shifted into deeper layers.[1] This clear pattern is consistent with later layers capturing more global document-wide information. Yet, whether this coding should be seen as true abstraction is less obvious, as the classification reflects just basic information from only the first sentence. Further experiments would dig deeper into this idea using a much larger model and produce more peculiar deviations from an abstraction hierarchy conception.

### Experiment 5: Large llama

**Analyses 1**

An extended last experiment probed a larger model: Llama-3.3-70b-Instruct, which contains 80 layers; Llama-3.3, which only provides an "Instruct" variant, as of December 2024.[2] This larger model was submitted to several of the same experiments as before (**Figure 5**). In terms of classification accuracy, the larger model performed about as well as the smaller one for the easier classification tasks but achieved better accuracy in the difficult ones. That is, the larger model yielded similar high points in accuracy for object-scene likelihood (**Figure 2a**) and object-object relatedness (**Figure 2b**). By contrast, the larger model accuracy was improved for animal/food classification (**Figure 2c**), which may reflect this being a less obvious property than generic relatedness. Larger model accuracy was also greater for the analogy-hard classification (**Figure 3c**), which may reflect the difficulty. Additionally, the larger model found greater success for each of the buried-concept classifications, which is consistent with generally improved long-context capabilities.

Along with these sensible improvements in accuracy, there was also one more surprising glaring pattern, which emerged for the two-item relation and four-item analogy experiments: All four experiments show two robustly distinct peaks, which are most evident in the attention and FFN layer outputs (**Figures 5c-f**). The first peak emerges around layers 12-16 and the second around layers 25-33. In other words, these semantic properties are most represented in two distinct spots. In between these peaks, there is a dip. If the peaks are taken to shed light on abstraction, this may reflect an increase of abstraction (producing the first peak then dip), then a decrease (producing the second peak), and then a further increase in abstraction (producing the second-half descent). None of the experiments on the smaller Llama-3.2-3b produced double peaks suggesting that this is an emergent property of scaling. Next, analyses turned to understanding what the model is accomplishing through this valley.

---

[1] Adding filler before the target text (e.g., "… explore. An {Item 1} and {Item 2}") had virtually no effect on SVM accuracies.

[2] The Instruct version of the model was used as this was the only variant provided as Llama-3.3, the most recent version in the Llama line of models, as of December 2024.

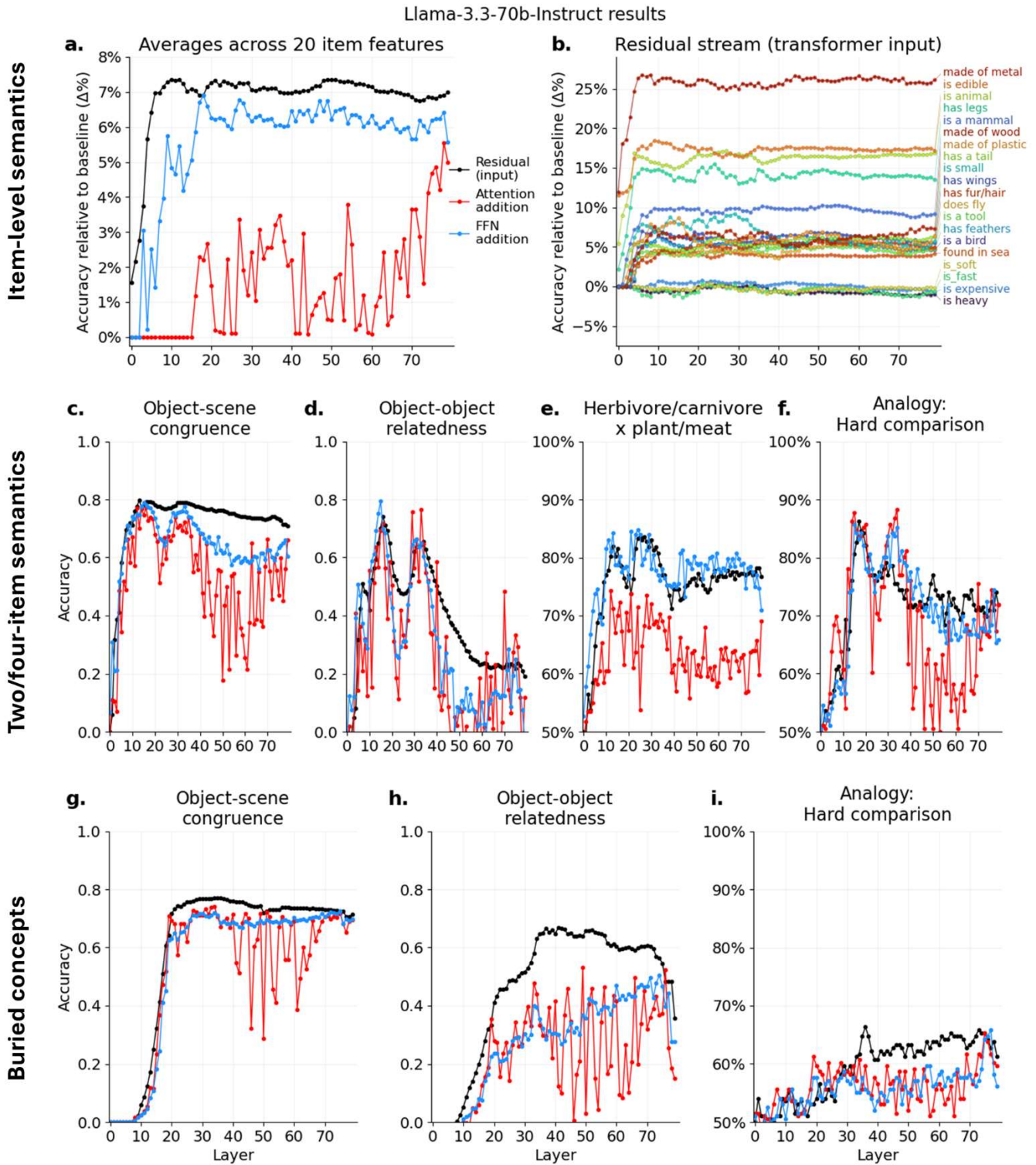

*Figure 5. Buried information representation.* Several but not all of the earlier experiments were performed again, now using Llama-3.3-70b-Instruct rather than Llama-3.2-3b. Each subfigure here maps to one of the subfigures from Figures 1, 2, 3, or 4. The mappings are specifically: **a.** Figure 1a, **b.** Figure 1b, **c.** Figure 2a left, **d.** Figure 2b, **e.** Figure 2c, **f.** Figure 3c, **g.** Figure 4c, **h.** Figure 4d, and **i.** Figure 4h.

To better understand the dip-valley-dip representational profile, I examined the attention output representations more carefully, as this component showed the largest swings. For this component, all seven of the test experiments' accuracies were z-scored (across the different layers for a given experiment). Then, the z-scored accuracies were overlaid, which is shown in **Figure 6a**. Remarkably, the valley in the two/four-item experiments coincides with a rise in representation seen in the buried-concept experiments (the rising blue lines in **Figure 6a**).

This is consistent with a steady rise in abstraction through the start of the valley. The overlay also shows how the buried concepts' peak representation overlays with the second peak of the non-buried concept; this is also evident looking back to **Figures 5g-i**, where the residual stream's representation of buried concepts is strongest around layer 35.

It is unclear what interpretation can unify these results. Potentially, there exist two functional hierarchies: the first hierarchy processes information from syntax to roughly sentence/fragment-level semantics, and this hierarchy's representations inform a growing second hierarchy that can capture a more global context. Alternatively, the valley may occur because the model referencing is global information to update the local representations initially produced at the first peak. However, to be clear, there are speculations, and further work is necessary to parse the mechanisms at play.

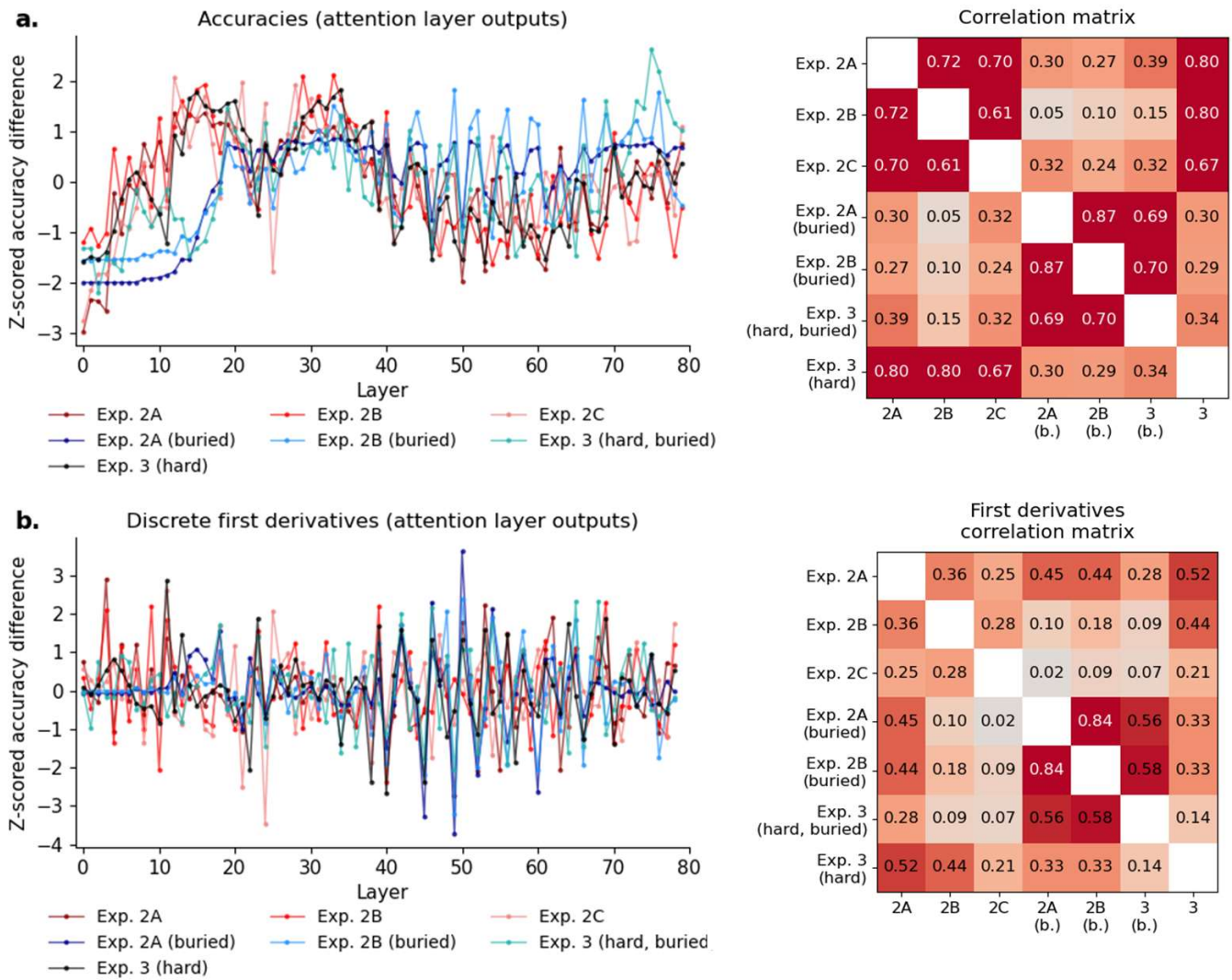

***Figure 6. Parallels between model accuracies and their accuracy first derivatives. a.*** *These series constitute the z-score of the attention output accuracies taken from Figures 5c-i; z-scored within-series. The correlation matrix was computed with respect to these 80-element-long series.* ***b.*** *The multiple series here are the discrete first derivative of attention output accuracies. The correlation matrix reflects just correlations between the second halves of the series as these are meant to hone into their anti-persistent (zigzag) portions.*

**Analyses 2**

One interesting final pattern noticed in the large model data is the large fluctuation in the back halves of the attention-layer outputs (notice zigzags in layers 40-70 of **Figures 5c-i**). That is, the information that the attention layers add to the residual stream shifts each layer between strongly representing the two/four-item semantic information to weakly representing said information. This can be quantified using the accuracy series' first derivatives – i.e., the differences between neighboring layers' accuracies. **Figure 6b** shows these first derivatives and the prominent fluctuations. This can be quantified as the first derivative's (Spearman) autocorrelation, and this is consistently negative (mean: $\rho = -.33$ [standard deviation = .12]), which is a phenomenon referred to as anti-persistence. This is specifically an effect between directly adjacent layers, as the autocorrelations between the accuracy difference of layers two, three, or four layers apart are all near zero ($-.08 < \rho s < .09$). However, studying a yet larger model, such as the 405b-parameter variant of Llama, could produce multi-layer patterns. Regardless, this anti-persistence is a curious pattern.

Adding further peculiarity, the precise high/low layers in this fluctuation are largely consistent across experiments: Notice the first-derivative accuracy overlap across experiments in **Figure 6b** (**left**) along with the strong correlations between experiments in **Figure 6b** (**right**) (correlation matrix mean: $\rho = .31$ [SD = .20]). This consistency suggests that thy e attention layers are not repeatedly overshooting some desired spot in latent space, but rather there is a type of coordination between adjacent layers.[3] To a degree, these fluctuations seem to be emergent properties of scaling. Examining Llama-3.2-3b's attention output accuracies also shows anti-persistence (mean: $\rho = -.41$ [SD = .15]), but these are not consistent across experiments (correlation matrix mean: $\rho = .06$ [SD = .20]). Returning to the initial focus on functional abstraction hierarchies, these fluctuations are not necessarily inconsistent with that idea. However, these fluctuations represent a separate aspect of LLM's functional architectures.

**Conclusion**

This research advances our understanding of how transformer layers encode semantic information, revealing both expected hierarchical trends and deviations from these trends.[4] These findings refine prevailing assumptions about transformer dynamics and open avenues for exploring how architectural innovations and scaling influence the emergent properties of modern LLMs.

---

[3] If only layer 40 and deeper are examined, FFN output accuracy also shows anti-persistence (mean $\rho = -.34$ [SD = .14]). However, the absolute levels of the FFN layer fluctuations are much smaller than those of the attention layer (see **Figures 5c-i**), and the FFN outputs show less consistency across experiments (correlation matrix mean $\rho = .18$ [SD = .16]; matrix not shown).

[4] Some tests were also conducted on grammar/tense/syntax. Representations distinguishing texts with the past vs. present vs. future tense appeared to peak early in layers 3-5. Additionally, analyses on grammatically proper versus improper (e.g., "I are" or "I been have") texts also showed an early peak (layers 3 & 4). However, analyses on simple vs. perfect tense showed representation rising steadily until layer 14 and staying maximal across the remaining layers. Thus, the results on basic grammar were inconclusive and not reported.

One of the novel components of the present research was the investigation of how "buried" concepts are represented in Experiments 5 and 6. For this, the input text consisted of the target of the research briefly (e.g., "An {item 1} and {item 2}"), followed by ~100 words of filler that is constant across every example. Then, analyses examined whether a model's activations after parsing the filter still represent the target despite the addition filler – i.e., asking whether the target representation persists. This was the case, but representation shifted starkly toward later layers. Potentially, this reflects a general compression of all information from past tokens, even if the past tokens are entirely unrelated to future ones. If so, deeper layers are not only identifying abstractions but rather, depth enhances the window of content processed even if there is little form to the information being compressed.

These results on buried concepts suggest that manipulating activations deep in a layer can simulate noticing a word in the past. This may be useful for steering and may open the door to "analogical steering", where the goal is to loosely encourage a model to approach some concept. For instance, prior work has shown how middle-layer representations of a concept can be identified by examining the activation patterns associated with the most recent concept (e.g., the activation patterns of the "Golden Gate Bridge").[7] By stimulating the middle layer components activated by the concept, researchers can cause a model's outputs to sharply discuss said concept (e.g., have the model discuss the Golden Gate Bridge). However, the present work shows how late layer representation of "buried" concepts can be extracted – i.e., the representation of the "Golden Gate Bridge" mentioned 100 filler words in the past. Steering as so may allow aspects of the representation to color the model's output without causing a fixation on the concept itself. This type of late-layer steering may allow a model to be gently nudged along a long-context window. In addition, this type of steering may be useful for implementing broadly relevant safety-related goals.

The large model (larger Llama-3.3-70b-Instruct) showed one stark deviation from a steady abstraction hierarchy. For the representation of both two-item relations and four-item analogies in simple texts, the model displayed two distinct and consistent representation peaks in its middle layers. That is, the representation of this local information first peaked in layers 12-16, then dipped, and then rose again and displayed a second peak in layers 25-33. Analysis of the buried concepts helped unpack this phenomenon, showing how representation of two/four-item relational information historic in the context window peaked around layer 33. The double-peak pattern points to how scaling introduces new emergent behaviors. While this may enhance model capabilities, it also complicates interpretability, making it harder to anticipate or control model outputs.

Emergent deviations from functional hierarchies in LLMs 13